\documentclass[accepted]{uai2023} 

\usepackage[american]{babel}

\usepackage{natbib} 
\usepackage{mathtools} 
\usepackage{microtype}
\usepackage{graphicx}
\usepackage{subfigure}
\usepackage{booktabs} 

\usepackage[utf8]{inputenc} 
\usepackage[T1]{fontenc}    
\usepackage{hyperref}       
\usepackage{url}            
\usepackage{booktabs}       
\usepackage{amsfonts}       
\usepackage{nicefrac}       
\usepackage{microtype}      
\usepackage{xcolor}         

\usepackage{amsmath}
\usepackage{amssymb}
\usepackage{mathtools}
\usepackage{amsthm}
\theoremstyle{definition}
\newtheorem{definition}{Definition}[section]

\usepackage{times}
\usepackage{soul}
\usepackage{url}
\usepackage[utf8]{inputenc}
\usepackage{multirow}
\usepackage{algorithm}
\usepackage{algorithmic}
\usepackage{wrapfig}

\title{Differentiable User Models}

\author[1]{\href{mailto:alex.hamalainen@aalto.fi?Subject=Your UAI 2023 paper}{Alex Hämäläinen}{}}
\author[1]{Mustafa Mert Çelikok}
\author[1,2]{Samuel Kaski}
\affil[1]{%
    Department of Computer Science\\
    Aalto University
}
\affil[2]{%
    Department of Computer Science\\
    University of Manchester
}
  
\begin{document}
\maketitle

\begin{abstract}
    Probabilistic user modeling is essential for building machine learning systems in the ubiquitous cases with humans in the loop. However, modern advanced user models, often designed as cognitive behavior simulators, are incompatible with modern machine learning pipelines and computationally prohibitive for most practical applications. We address this problem by introducing widely-applicable differentiable surrogates for bypassing this computational bottleneck; the surrogates enable computationally efficient inference with modern cognitive models. We show experimentally that modeling capabilities comparable to the only available solution, existing likelihood-free inference methods, are achievable with a computational cost suitable for online applications. Finally, we demonstrate how AI-assistants can now use cognitive models for online interaction in a menu-search task, which has so far required hours of computation during interaction.
\end{abstract}

\section{Introduction} \label{introduction}

User modeling constructs informative representations of individual users to enable computational systems to customize and adapt their behavior for them \citep{li2020survey}. It has been extensively studied over the years, also recently in recommender systems \citep{yu2019adaptive, yuan2020parameter}, human-in-the-loop machine learning \citep{daee2018user} and AI-assistants \citep{horvitz2013lumiere, dafoe2021cooperative}. Machine learning is needed in user modeling to infer user-specific information based on observed user behavior. Depending on the application, the inferred information can be the end result or, for instance, used to parameterize a user simulator to form predictions of user behaviors to guide the behavior of the system.

Traditionally, salient use cases of user modeling, such as recommendation engines, have utilized user-specific preference profiles based on users' history. However, these approaches are not sufficiently powerful in more complex interactive applications, where the user plans and interacts strategically, for instance in human-AI collaboration and human-in-the-loop decision-making. On the other hand, while recent ML research has shown significant success in learning accurate neural models directly from data, this is an infeasible approach in user modeling in general, as typical user-driven applications are data starved. In contrast, recent approaches, e.g., \cite{kangasraasio2019parameter, moon2022speeding}, have utilized advanced general-purpose behavioral models, based on cognitive science, in a Bayesian setting and received encouraging results with limited data. The probabilistic treatment of the problem enables taking uncertainty of the inferences into account --- which fundamentally allows the system to balance between exploration and exploitation in interaction with the user.

A prominent body of these advanced cognitive models is based on computational rationality \citep{lewis2014computational, gershman2015computational}, which posits that seemingly irrational behaviors of humans are rational under their cognitive bounds. It follows that human behaviors can be accurately modeled as a result of RL-based optimization given that the underlying decision-theoretic framework and the optimization procedure are specified such that they accurately capture the appropriate bounds governing human cognition, such as the limits of computational capacity. A concrete example of such a model, which we consider in our experiments, is the model of rational menu search \citep{chen2015emergence}. This cognitive model describes human search behavior in terms of eye movements when searching for a target item in a computer dropdown menu, while encoding the limitations of human cognition and perception when processing visual information.


Despite their many benefits, these advanced cognitive models are currently not used beyond small-scale practical applications due to two important factors: (1) they are expressed as non-differentiable simulators and hence incompatible with modern machine learning frameworks and (2) they are computationally infeasible to be used directly in realistic applications.

In this paper, we address these limitations: we enable computationally efficient probabilistic user modeling suitable for real-time applications --- even with advanced cognitive models that lack a closed-form likelihood. We do this by combining the best of gradient-based and Bayesian learning: we show how one can develop differentiable user models which are sample-efficient by leveraging prior knowledge from non-differentiable cognitive models and can quantify uncertainty in their estimates. As a result, the surrogates become widely applicable with online computational cost independent of the complexity of the original models. The contributions of this work are: \begin{itemize}
    \item We introduce a way of enabling computationally efficient inference with cognitive user models by building generalizable differentiable surrogates for them through meta-learning.
    \item We demonstrate a flexible way of leveraging any existing user data during surrogate training to address possible model misspecification in cognitive models, especially in the case of action noise.  
    \item With neural processes as example surrogate models, we demonstrate comparable user modeling accuracy to current methods with a computational time suitable for online applications.
\end{itemize}
Our work removes a key computational bottleneck currently hindering incorporation of users into probabilistic programming models. Probabilistic user modeling based on cognitive models can now be applied widely without extensive computational budgets.

\section{Differentiable user models} \label{dum}

This work considers computationally efficient probabilistic user modeling for interactive settings, between a user from a user population and a computational system. User modeling is brought in to guide adaptation of the behavior of such a system for individual users; user modeling is needed for (i) inferring user-specific information from observed user behaviors and (ii) then using the information for user behavior prediction. The following subsections detail the specifics of current approaches and their limitations, together with our approach for addressing these limitations.

\subsection{Probabilistic user modeling with cognitive models}

Following the intuition presented by \cite{kangasraasio2019parameter}, we formulate the probabilistic user modeling setting as follows: a population of users is engaged with a distribution of decision-making tasks denoted by $p(\theta_T)$. Each modeling scenario involves a user $\theta_U \sim p(\theta_U)$ and is fully described by the respective user and task specific parameters $\theta = \{\theta_T, \theta_U\}$. The users are assumed to generate their policies $\pi$ through an implicit process $\mathcal{P}_\theta$ which they execute to generate pairs of states and actions $(\mathbf{s}, \mathbf{a})$. The system has access to a cognitive model $p(\pi \mid \theta)$ approximating the true process and a corresponding prior $p(\theta)$. An important task corresponding to this user modeling setting is the inference problem of approximating the posterior 
\begin{equation}
    p(\theta \mid (\mathbf{s}, \mathbf{a})) \propto p((\mathbf{s}, \mathbf{a}) \mid \pi)p(\pi \mid \theta) p(\theta).
\end{equation}
Computing the posterior $p(\theta \mid (\mathbf{s}, \mathbf{a}))$ and then using the likelihood model $p(\pi \mid \theta)$ for computing the corresponding posterior predictive distribution $p(\pi \mid (\mathbf{s}, \mathbf{a})) = \int p(\pi \mid \theta) p(\theta \mid (\mathbf{s}, \mathbf{a})) d\theta$ would be the Bayesian choices for achieving the objectives (i) and (ii). 






For cognitive models, the likelihood $p(\pi \mid \theta)$, required for solving the Bayesian inference task for the posterior, is typically not evaluable in closed-form due to the simulator-type nature of these models. So far, this issue has been circumvented by utilizing exclusively likelihood-free inference (LFI) methods, such as approximate Bayesian computation (ABC) \citep{sisson2018handbook, sunnaaker2013approximate} and Bayesian optimization for likelihood-free inference (BOLFI) \citep{gutmann2016bayesian}, as proposed by \citet{kangasraasio2019parameter} and \citet{moon2022speeding}. The basic idea of LFI is to replace the computationally expensive simulator $p(\pi \mid \theta)$ with an approximation that is separately learned on the observed data \citep{gutmann2016bayesian}, in this case the $(\mathbf{s}, \mathbf{a})$ from each user. For user modeling, this approach has two problems: this process requires numerous computationally expensive simulations with the cognitive model and the data in typical user modeling applications often is too scarce for learning a new model independently for each user.

\cite{moon2022speeding} proposed circumventing the computational complexity by learning a generalizable policy-modulation network as a surrogate for the original model, i.e. $p(\pi \mid \theta)$, and obtained significant speed-ups for inference. However, as noted by the authors, their approach is still prohibited by the computational cost of LFI needed for approximating the posterior, and is not feasible for real-time inference. Similarly, the simulation costs of cognitive models $p(\pi \mid \theta)$ are often too expensive to enable estimating the posterior predictive in real-time applications, even if the posterior was readily available. Furthermore, even though LFI methods are developing fast, practical interactive settings may require hierarchical approaches for generalizing across the user models, which has been traditionally difficult with LFI \citep{turner2014hierarchical}. An additional problem with LFI-based modeling is the sensitivity to model misspecification, which is very likely in user models.


\subsection{Amortization for cognitive models}

In this work, we seek to address the limitations of current approaches and to enable efficient computation for both the likelihood and posterior models so that the posterior predictive distribution is practical to approximate. Our approach is to amortize posterior predictive inference through surrogate modeling. Training generalizable surrogates offline enables using them during online interaction without extensive computation.

While this approach would solve the issue of online computational complexity, the offline complexity of simulating sufficient amounts of training data for them will still be an issue due to the vast diversity of different behaviors the cognitive models are able to express. In particular, even if training a generalizable surrogate for a cognitive simulator would be achievable, as done in the work of \cite{moon2022speeding}, training a surrogate directly for approximating the posterior can be ultimately be computationally intractable if constructing the training data requires numerous repeated evaluations with LFI (for the reference, \citet{kangasraasio2019parameter} reported that even a single LFI result would require at least 700 CPU hours with the menu search model). Furthermore, as we will later discuss in Section~\ref{ipum}, data-efficiency in training the surrogates is also otherwise a desirable factor as it helps combating model misspecification in cognitive models.

\subsection{Casting simulator-based modeling as meta-learning}

In order to make the surrogate training more sample-efficient, we approach amortization task from meta-learning perspective. Here, the key insight is that both the likelihood and posterior models can be learned jointly with an appropriate policy approximation task, without ever needing to approximate the true posterior $p(\theta \mid (\mathbf{s}, \mathbf{a}))$, if one is satisfied with using a latent representation $z \in \mathcal{Z}$ to capture user-specific information. Following this intuition, we generalize the likelihood and posterior models to mappings $h$ and $g$:
\begin{definition}[Amortization for cognitive models] \label{def}
Let $\mathcal{S}$ and $\mathcal{A}$ denote the state and action spaces corresponding to the user model and $\mathcal{O} = \bigcup_n (\mathcal{S} \times \mathcal{A})^n$ be a collection of $m$ observations over behavior generated by an individual user $\theta_U$ in a task $\theta_T$. Amortization for cognitive models corresponds learning the following functions such that they are evaluable during online interaction: \begin{enumerate}
    \item Inference of user and task representations, done by the mapping $h:~\mathcal{O} \rightarrow P(\mathcal{Z})$, where $P(\mathcal{Z})$ denotes a probability distribution over a joint user and task representation space $\mathcal{Z}$, which aims to capture the properties governing user behavior.
    \item User behavior prediction, done by the mapping $g:~\mathcal{S} \times \mathcal{Z} \rightarrow P(\mathcal{A})$, where $P(\mathcal{A})$ is a probability distribution over user action space.
\end{enumerate}
\end{definition}

In line with the Definition~\ref{def}, we amortize the computation for the posterior predictive distribution over a cognitive model through learning generalizable surrogates for the mappings $h$ and $g$. Intuitively, we are here building on the conceptual similarity between Bayesian methods and meta-learning (previously discussed, e.g., by \citet{grant2018recasting} and \citet{garnelo2018neural}), and consider the mapping $h$, i.e., computing the posterior over the user representation as equivalent to task-specific adaptation during meta-testing and the mapping $g$, i.e., computing the likelihood as analogous to prediction.


To formalize the idea, let $s \in \mathcal{S}$, $a \in \mathcal{A}$ and $(\mathbf{s}, \mathbf{a}) = \{(s_1, a_1), \dots, (s_n, a_n)\} \in \mathcal{O}$. Our goal is to learn the mappings $h$ and $g$, with optimizable parameters $\{\psi, \phi\}$, to approximate the posterior $p_\psi(z \mid (\mathbf{s}, \mathbf{a}))$ and the likelihood $p_\phi(a \mid s, z)$ with respect to a latent representation $z \in \mathcal{Z}$. The corresponding posterior predictive model can here be written as $p_{\{\psi, \phi\}}(a \mid s, (\mathbf{s}, \mathbf{a})) = \int p_\phi(a \mid s, z) p_\psi(z \mid (\mathbf{s}, \mathbf{a})) dz$. The surrogates should jointly minimize the following objective for policy approximation, while generalizing over the ground-truth user and task population ($\theta \sim p(\theta)$ and $\pi \sim p(\pi \mid \theta)$):
\begin{equation}
    \min_{\phi, \psi} \mathbb{E}_{\theta \sim p(\theta), s \in \mathcal{S}} \bigg[ \delta \big[ \pi(a \mid s), p_{\{\psi, \phi\}}(a \mid s, (\mathbf{s}, \mathbf{a})) \big] \bigg],
\label{eq:1}
\end{equation}
where $\delta$ is a dissimilarity function (e.g., KL-divergence) and the observations $(\mathbf{s}, \mathbf{a})$ are assumed to have been generated by executing $\pi$ in the underlying environment. In Section~\ref{nps}, we demonstrate how a solution to this problem can be approximated with neural processes.




Consistently with numerous current meta-learning approaches (e.g., \citet{finn2017model}, \citet{garnelo2018neural}), we propose a modeling workflow with separate offline (meta-training) and online (meta-testing) phases described below. We additionally expand on mitigating the effects of possible model misspecification in cognitive models.

\subsection{Meta-training and meta-testing} Algorithm~\ref{alg:example} specifies the proposed meta-training procedure, to enable generalization of the surrogates $h$ and $g$ over the population of interest $p(\theta)$. The procedure needs to be complemented with an appropriate meta-learning loss for approximating a solution to Eqn.~\ref{eq:1} in terms of $\{\psi, \phi\}$, depending on the implementations of $h$ and $g$. In Section~\ref{nps}, we exemplify this with neural processes.

The corresponding meta-testing, i.e., task-specific adaptation phase is straightforward: mappings $h$ and $g$ can be utilized for inferring user representations $z \sim p_\psi(z \mid (\mathbf{s}, \mathbf{a}))$ w.r.t. observed $(\mathbf{s}, \mathbf{a})$ and for predicting user behaviors $a \sim p_\phi(a \mid s, z)$ on states of interest $s \in \mathcal{S}$.
\begin{algorithm}[]
   \caption{Meta-training cognitive model surrogates}
   \label{alg:example}
\begin{algorithmic}
   \STATE {\bfseries Require:} A distribution over users: $p(\theta_U)$
   \STATE {\bfseries Require:} A distribution over tasks: $p(\theta_T)$
   \STATE {\bfseries Require:} A cognitive model: $p(\pi \mid \theta)$
   \STATE Initialize $h$ and $g$ with $\{\psi, \phi\}$
   \REPEAT
   \STATE Sample $\theta = \{\theta_U, \theta_T\}$, $\theta_U \sim p(\theta_U)$, $\theta_T \sim p(\theta_T)$
   \STATE Generate $\pi \sim p(\pi \mid \theta)$
   \STATE Generate $n$ trajectories $(\mathbf{s}, \mathbf{a})$ by executing $\pi$
   \STATE Optimize $\{\psi, \phi\}$ with respect to $(\mathbf{s}, \mathbf{a})$ with an appropriate training loss
   \UNTIL{done}
\end{algorithmic}
\end{algorithm}

Note that consistently with \cite{garnelo2018neural}, the proposed meta-learning workflow deliberately differs from many other popular meta-learning approaches, such as model-agnostic meta-learning (MAML) \citep{finn2017model} and Reptile \citep{nichol2018reptile}, by fully excluding the gradient-based optimization loop during task-specific adaptation phase. Instead, the adaptation phase is here reduced to a forward pass through $h$. Not only is this computationally faster, enabling online computation, the probabilistic nature of our approach can also enable interactive systems to balance between exploration-exploitation trade-offs. As we demonstrate in our experiments, these benefits additionally translate into improved modeling accuracy.



\subsection{Model misspecification in cognitive models.} \label{ipum}

Model misspecification is a relevant issue in behavioral user modeling. While typical LFI-approaches are highly sensitive to misspecification, this can be mitigated with our approach by combining observed user data with simulated data and meta-training the surrogates again, when new observations become available. We demonstrate in Section~\ref{ms} that this approach enables balancing between modeling accuracy and data requirements --- especially in practical interactive user modeling applications which only have limited collections of user behavior datasets available. 

\section{User modeling with neural processes} \label{nps}

We use neural processes (NP) \citep{garnelo2018neural} as an example solution for implementing and learning the mappings $h$ and $g$ of Definition~\ref{def}. First, we briefly cover the relevant background on NPs and then explain in detail how they can be adapted for user modeling.

\subsection{Background on neural processes}

Neural processes \citep{garnelo2018neural} are a family of neural latent-variable models combining properties of neural networks and Gaussian processes (GP). Specifically, they are differentiable solutions for representing uncertainty over functions that may be utilized for few-shot approximation. For our purposes, NPs are particularly fitting as they match Definition~\ref{def} and that the meta-learning objective (Eqn.~\ref{eq:1}) can be readily computed for them.


NPs model a set of functions $\{f_d\}_d$ where each $f_d: X \rightarrow Y$ is assumed to be drawn from an underlying stochastic process $f_d \sim F$. NP approximates the underlying process $F$ with a neural network $g$. As each function $f_d$ drawn from the process $F$ represents an individual instantiation of the process, a latent variable $z$ is introduced for capturing the instance-dependent variation in $F$ as $f_d(x) = g(x,z)$. 
NPs consist of an encoder, an aggregator and a conditional decoder. The encoder is a neural network for constructing representations $r_i = h_\phi((x,y)_i)$ at given observations $(x,y)_i$. The aggregator, $\alpha$, constructs permutation-invariant summaries of the encoded representations as $r = \alpha(\{r_i\}) = \frac{1}{n}\sum_{i=1}^n r_i$. The summaries are further utilized to parametrize a (multivariate Gaussian) latent distribution $z \sim \mathcal{N}(\mu(r), I\sigma(r))$. The conditional decoder, $g_\psi(x_T, z)$, is a neural network that is conditioned on samples from the latent distribution to estimate $f_d(x_T) = y$ at locations $x_T$.

NP meta-training procedure samples individual instantiations $f_d \sim F$ of the stochastic process $F$. Here, each function $f_d$ is evaluated at a varying number of inputs to produce a dataset of tuples $(x, y)_i^d$. Each dataset is then divided into separate \textit{context} $(x_{1:m},y_{1:m})$ and \textit{target} $(x_{m+1:n}, y_{m+1:n})$ sets. Intuitively, here the context set represents the fully observed function evaluations while the target $x_{m+1:n}$ represents the locations at which the model aims to approximate $y_{m+1:n}$. The context and target sets are input to the encoder and the conditional decoder respectively, and the model parameters $\{\phi, \psi\}$ are optimized with respect to the NP-variant of Evidence lower-bound (ELBO). For further information about NPs and their training, see \cite{garnelo2018neural}. Finally, note that the low complexity of NPs ($\mathcal{O}(n+m)$) makes them suitable for real-time scenarios.

\subsection{Adapting neural processes for user modeling}

Neural processes can be adapted as concrete implementations for the required mappings $h$ and $g$ and for approximating a solution for Equation~\ref{eq:1} within the proposed meta-training procedure (Algorithm~\ref{alg:example}). First, we recognize that Equation~\ref{eq:1} is essentially a function approximation problem to which NPs can be applied --- the true behavior-generative process $p(\pi \mid \theta)$ can essentially be treated as a stochastic process $\mathcal{P}$ where each instantiation $\pi \sim \mathcal{P}$ represents a policy. The NP latent variable $z$ is utilized for capturing user/task representations and the mappings $h$ and $g$ can be implemented with the NP encoder $p_\phi(\pi \mid z)$ and conditional decoder $p_\psi(z \mid (\mathbf{s}, \mathbf{a}))$. The meta-training procedure is adapted as follows: the sampled behaviors $(\mathbf{s}, \mathbf{a})$ are split into context and target sets and the parameters $\{\psi, \phi\}$ can be optimized according to NP-ELBO.

In addition to the vanilla NPs, we consider also attentive neural processes (ANP) \cite{kim2019attentive}, conditional neural processes (CNP) \citep{garnelo2018conditional} and attentive conditional neural processes (ACNP). ANPs are essentially NPs with the difference of including attention in the encoder architecture. The attention acts as a local latent variable, allowing ANPs to capture both global and local information affecting user behaviors. CNPs (and ACNPs) implement the latent encoding $h$ as a deterministic mapping, thus lacking an important ability of sampling on $\mathcal{Z}$.

\section{Experiments}

We conduct three experiments where we compare our approach against other ways one could conceivably try to solve the problem --- although this has not been previously done. The first is a demonstration in a benchmark gridworld environment. The second is a menu search task where a cognitive user model, justified and validated by earlier cognitive science studies, allows us to study real-user performance with simulations. The third experiment is a reasonably realistic menu search assistant scenario.

\paragraph{Comparison methods and baselines.} We assess the modeling capabilities of the proposed solution in terms of its ability to predict the actions of individual agents, as a function of the number of previous observations of their behavior in the modeling task of Equation~\ref{eq:1}. This metric directly evaluates the posterior predictive but also indirectly the quality of the posteriors over user representations $z \in \mathcal{Z}$. Unless otherwise specified, the experiments aim to simulate realistic user modeling applications by limiting the training data to observations from $\sim 1000$ simulated users.

\begin{table*}[]
    \centering
    
    \caption{Modeling accuracy as a function of the number of observed full episodes in the menu-search setting of Section~\ref{ms}.}
    
    \begin{tabular}{r|c|c|c|c|c|c}
        Episodes & ANP & Reptile & MAML & Transformer & Oracle & Population avg. \\ \hline
        $0$ & $\mathbf{0.937 \pm 0.011}$ & $0.829 \pm 0.054$ & $0.774 \pm 0.033$ & $0.920 \pm 0.035$ & $\mathbf{0.970 \pm 0.002}$ & $0.921 \pm 0.012$ \\
        $1$ & $\mathbf{0.953 \pm 0.011}$ & $0.921 \pm 0.021$ & $0.916 \pm 0.026$ & $0.922 \pm 0.021$ & $\dots$ & $\dots$ \\
        $2$ & $\mathbf{0.954 \pm 0.011}$ & $0.930 \pm 0.020$ & $0.928 \pm 0.025$ & $0.931 \pm 0.017$ & $\dots$ & $\dots$ \\
        $5$ & $\mathbf{0.955 \pm 0.010}$ & $0.944 \pm 0.017$ & $0.943 \pm 0.021$ & $0.926 \pm 0.012$ & $\dots$ & $\dots$ \\
        $9$ & $\mathbf{0.955 \pm 0.010}$ & $0.954 \pm 0.016$ & $0.952 \pm 0.014$ & $0.928 \pm 0.009$ & $\dots$ & $\dots$ \\
    \end{tabular}
    \label{tab:ex2tab}
\end{table*}

We compare our approach against two baselines and three alternative surrogate architectures. The alternative surrogates are transformers trained with MAML and Reptile, and a standard transformer. MAML and Reptile act as alternative representative meta-learning approaches to the policy approximation task over user population, while the transformer intends to provide a reference point for the performance of sequential models which are frequently used in alternative user modeling domains, such as sequential recommendation. None of the alternative surrogate architectures are fully consistent with the proposed meta-learning procedure and are applied to the policy approximation task on simulated data directly instead. We also include comparisons between several alternative NP architectures. Details are included in the Supplement. 


\begin{figure}[]
\begin{subfigure}
    \centering
    \includegraphics[width=0.48\linewidth]{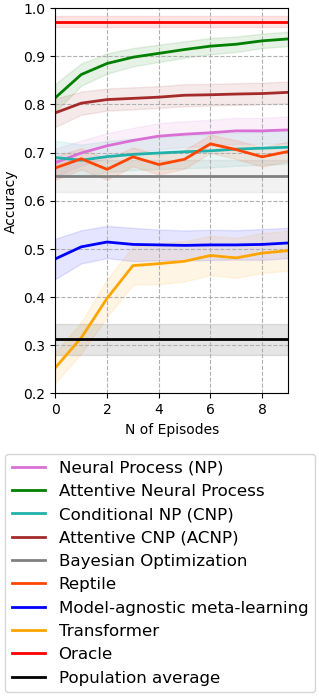}
\end{subfigure}
\hfill
\begin{subfigure}
    \centering
    \includegraphics[width=0.46\linewidth, trim = {0.0cm, -1.9cm, 0.0cm, -0.0cm}]{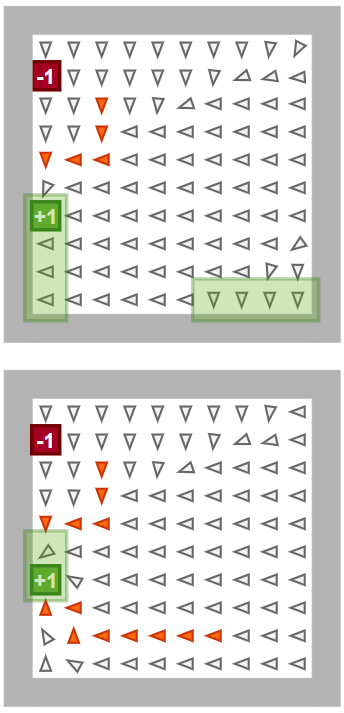}
\end{subfigure}

\caption{Gridworld results. \textbf{Left:} Modeling accuracy as a function of the number of observed full episodes. The best NP-based model (here ANP) achieves comparable results to the upper bound given by an oracle; all NP-based models are clearly better than alternatives. The figure illustrates the gradual improvement of the predictions of NPs as more episodes are perceived. The BO results are averaged over the number of context trajectories due to the small sample size. \textbf{Right:} Illustration of ANP uncertainty update on policy predictions. The predictions (gray arrows) align towards the implicitly inferred possible goal states (green rectangles). In the upper figure, the predictions are conditioned on one observed trajectory (orange arrows). In the lower figure, we observe that the system implicitly infers the location of the positive reward, within the accuracy of two squares, after perceiving the second trajectory. The dark green and red squares are the true positive and negative reward states.}

\label{fig:ex1_1}
\end{figure}

The two baselines are a Bayesian Optimization (BO) model and a population average predictor. Furthermore, we provide results from an oracle, acting as an upper bound for the performance of any solution, including LFI. Both the BO baseline and the oracle utilize the cognitive model $p(\pi \mid \theta)$ directly for prediction --- the oracle parametrizes the cognitive model with the true user parameters while BO utilizes MAP estimates. The population average predictor directly approximates the population level action distribution $p(a \mid s)$ for each state without any user-specific conditioning.

Even though it would be an important baseline, we were not able to produce any representative results with LFI due to its immense computational complexity. For reference, \cite{kangasraasio2019parameter} compared several LFI-methods, including BO, on exactly the same Menu Search model used in our experiments. They gave 700h of CPU-time for each method to run only one individual inference task and noted that it is likely that none of the methods converged. Obtaining conclusive accuracy results with LFI in our experiment setting is practically intractable as at least hundreds of individual inference results would be needed. Here, we consider the BO-baseline as an approximate lower-bound for LFI performance. Although converging faster than LFI, BO is still computationally very heavy, due to the expensive simulation costs, and feasible only in our first experiment.

\subsection{Experiment 1: Gridworld environment}

\paragraph{Setting.} The first experiment scenario is based on a simple $10 \times 10$ gridworld environment. In this setting, we consider modeling Monte Carlo Tree Search (MCTS) \citep{browne2012survey} agents with unknown reward functions and MCTS parameters. This benchmark scenario evaluates the modeling system's ability to approximate the uncertainty over user policies. In this experiment, we assume that a generative user model and a parameter prior are available, capturing the true generative process of the population.

The gridworld environment is defined as a partially observable Markov decision process (POMDP) with deterministic transition dynamics. The action space consists of four possible actions that correspond to the agent relocating from its current state to adjacent states. Each gridworld scenario always contains two reward states - one with a positive reward and one with a negative reward. The agents gain no rewards or penalties other than from the given states. The full setting details are given in the Supplement.

\paragraph{Modeling task.} The modeling task is to predict the subsequent actions of agents sampled from the population. Each scenario assigns the modeling system with observed trajectories from a varying number of previous episodes and a partial trajectory from the current episode generated by the agent. The task is to predict the remaining actions of the trajectory in the current episode. All information about the agent, excluding the observed trajectories, is hidden during both training and evaluation (except for the oracle).

\paragraph{Results.} The NP-family models are mostly able to outperform all the baselines (Fig.~\ref{fig:ex1_1}) with the ANP converging close to the performance of the oracle. It is likely that the BO-baseline has not properly converged, although given clearly the largest amount of computation time, and it may not act as a reliable approximation for LFI performance. Finally, the transformer and MAML are unable to generalize to the task, likely due to the too limited amount of training data.

Comparisons between the NP-family models suggest that, in terms of NP architecture, the most impactful factor contributing to the modeling performance is attention, i.e., local latent variables, as ANP and ACNP outperform their non-attentive counterparts. Consistent with the probabilistic treatment of $h$ (Def.~\ref{def}), stochasticity of the global latent variables $z$ (ANP and NP) also seems to improve the results. Because ANP was clearly the best of the NP methods, and hence remaining NP-models would not affect conclusions, we omit the NP, CNP and ACNP models for the following experiments, to save computation.

\subsection{Experiment 2: Menu search environment} \label{ms}

\paragraph{Setting.} Our second experiment is based on the Menu Search model of \citet{kangasraasio2019parameter}, a modified version of \citet{chen2015emergence}. The Menu Search model is a cognitive model describing human search behavior in terms of eye movements (saccades) when searching for a target item in a computer dropdown menu. Motivated by \textit{computational rationality} \citep{gershman2015computational}, the model simulates user behavior as a result of optimizing the search behavior with RL given their cognitive constraints of the user. The details are given in the Supplement.

\begin{table*}[]
    \centering
    
    \caption{Modeling accuracies for different numbers of observed full episodes with the ANP-based system when trained with data partially from a misspecified user model and partially from the true population. Here, the percentages denote the share of the training data generated with the misspecified user model.}
    
    \begin{tabular}{r|c|c|c|c|c}
        Episodes & ANP 0\% & ANP 25\% & ANP 50\% & ANP 75\% & ANP 100\% \\ \hline
        $0$ & $\mathbf{0.937 \pm 0.011}$ & $0.920 \pm 0.015$ & $0.895 \pm 0.016$ & $0.888 \pm 0.013$ & $0.852 \pm 0.017$ \\
        $1$ & $\mathbf{0.953 \pm 0.011}$ & $0.923 \pm 0.012$ & $0.899 \pm 0.014$ & $0.891 \pm 0.013$ & $0.854 \pm 0.016$ \\
        $2$ & $\mathbf{0.954 \pm 0.011}$ & $0.925 \pm 0.012$ & $0.901 \pm 0.014$ & $0.892 \pm 0.012$ & $0.857 \pm 0.016$ \\
        $5$ & $\mathbf{0.955 \pm 0.010}$ & $0.926 \pm 0.012$ & $0.902 \pm 0.014$ & $0.894 \pm 0.012$ & $0.862 \pm 0.016$ \\
        $9$ & $\mathbf{0.955 \pm 0.010}$ & $0.926 \pm 0.012$ & $0.902 \pm 0.014$ & $0.895 \pm 0.012$ & $0.865 \pm 0.016$ \\
    \end{tabular}
    \label{tab:ex2btab}
\end{table*}

\paragraph{Modeling task.} In this experiment, we apply our method for modeling agents whose search behavior is specified by the Menu Search model. As in the first experiment, we train the model parameters on data simulated with the given cognitive model. For each simulation, we sample a new menu, together with its element-wise information about the target word, as specified by \citet{kangasraasio2019parameter}.

%

\paragraph{Modeling accuracy.} Table~\ref{tab:ex2tab} summarizes the obtained modeling accuracies. We notice that after one observed trajectory, the ANP-based model achieves results comparable to the oracle upper bound. Unlike in the previous experiment, most of the users seemed to converge to a relatively narrow and finite set of search strategies, simplifying the difficulty of the modeling problem. As a result, MAML and the transformer achieve clearly higher relative accuracy than in the previous experiment, despite the limited training data.



\paragraph{Model misspecification.} We study the effects of model misspecification in cognitive models by repeating the modeling task with a noisy model. This model represents an otherwise accurate Menu Search model, but roughly $35\%$ of the saccades are modeled randomly into incorrect locations instead of following the policy of the correct model (full implementation details in the Supplement). We repeat the meta-training with different percentages of data obtained from the true user population. We explore both our solution's robustness against the model with action noise and its ability to adapt to the true generative process.

The results are gathered in Table~\ref{tab:ex2btab}. First, we observe that our approach can remain robust against user model noise: even when trained solely on data coming from the noisy model (ANP 100\%), the modeling accuracy remains reasonably good and surpasses the accuracy of the noisy model ($\approx65\%$). Secondly, it can be seen how our solution adapts to the ground-truth generative process when the proportion of the ground-truth data increases. We repeated the scenario by meta-training the ANP only on data from the true user population. We found that the noisy model improved the results when the number of observed real users was under $200$ (i.e., here percentage $<20\%$), after which it had a hindering effect on the predictions. However, we expect that the utility of misspecified models can be significantly higher in more complex modeling problems where more data is required to generalize to the problem.



\subsection{Experiment 3: Menu search assistant} \label{mse}

\paragraph{Setting.} In our third experiment, we aim to demonstrate the practical utility of the proposed approach for interactive systems by extending the Menu Search environment into a reasonably realistic AI-assistant scenario. First, we scale the environment to consider two levels of hierarchy: the menu consists of a main menu whose elements act as links to sub-menus; we use the menus of the previous experiment. 

Secondly, we introduce an AI assistant equipped with the proposed user modeling system. The task of the assistant is to utilize the modeling system to infer the target elements of the users based on observed search behaviors in the current menu, and to propose sub-menus for the users. Intuitively, a successful assistant should guide the users to menus that are likely to contain the true target for them, to shorten their search time. The assistant is allowed to provide any guidance only after the user is independently searched through at least one sub-menu. We implement the assistant as a simple rule-based agent that conditions its actions on the simulated user behaviors $a \sim p_\phi(a \mid s, z)$, $z \sim p_\psi(z \mid (\mathbf{s}, \mathbf{a}))$. Further details on the experiment setting are in the Supplement.

\begin{table*}[]
    \centering
    
    \caption{User search times and modeling/simulation times per assistant action with different assistant systems in Section~\ref{mse}.}
    
    \begin{tabular}{l|l|l|r}
        Assistant type & Search time (s) & Time saved (\%) & Modeling time per action (ms) \\ \hline
         No assistance        &  $4.774 \pm 0.235$ & $-$     & $- $ \\
         MAML                 &  $4.089 \pm 0.645$ & $14.3$   & $1174.922 \pm 43.760$ \\
         Reptile              &  $3.973 \pm 0.519$ & $16.8$   & $1053.342 \pm 37.988 $ \\
         Transformer          &  $2.918 \pm 0.191$ & $38.8$  & $\mathbf{1.140 \pm 0.442}$ \\
         ANP   &  $\mathbf{2.590 \pm 0.226}$ & $\mathbf{45.7}$  & $8.460 \pm 6.495$ \\
         Full knowledge       &  $\mathbf{2.577 \pm 0.162}$ & $\mathbf{46.0}$  & $- $ \\
    \end{tabular}
    \label{tab:ex1}
\end{table*}

\paragraph{Results.} Table~\ref{tab:ex1} compares the performance of the resulting assistant against a non-assisted user, a MAML-based, a Reptile-based, and a transformer-based solutions. The MAML and Reptile-based solutions require gradient-computation during test-time leading to modeling times greatly higher than the response time between human actions ($\approx 300$ms) in this experiment. This prevents online user model updates, hence hindering the effectiveness of the assistance. We also include results with an assistant that has full knowledge of the users' target elements to provide an upper-bound. We notice that the ANP-guided assistant can significantly reduce the user's search time and almost reaches the upper-bound performance of the assistant that has perfect knowledge. The observed results are encouraging regarding the ability of our solution to guide the behaviors of real-time interactive systems.

\section{Related work}

Our work connects to a larger body of research considering user modeling in interactive AI. For instance, \cite{carroll2019utility} and \cite{strouse2021collaborating} share the idea that efficient interaction with humans requires the AI to have an accurate model of the human. In contrast to many this line of works, our work concentrates on using models based on cognitive and behavioral sciences as priors, instead of ML-experts hand-crafting the models from scratch or learning them from large collections of user data. Using such models has been impractical up to now, and this the problem we now solve.

Inverse reinforcement learning (IRL) \citep{ng2000algorithms} considers a related problem to ours, aiming to recover agents' reward functions based on observed behaviors. Although it has been previously utilized also in user-centric problems \citep{chandramohan2011user}, our perspective is more general as we consider inference over arbitrary user parameters (instead of only rewards) and over varying policy-generative algorithms/processes. This allows our approach to be utilized for inference with a wide range cognitive models, where user behaviors are not necessarily optimal and are governed by human biases. Imitation learning (IL) \citep{hussein2017imitation}, on the other hand, considers learning models to imitate human (expert) behaviors on a given task. The crucial difference to our setting is that, unlike with IL, we do not necessarily seek to solve the task the human is solving, but to probabilistically model humans and their behaviors.

Using transfer and meta-learning in RL problems has been previously widely studied. For instance, \cite{yao2018direct} used HiP-MDPs \citep{doshi2016hidden} for modeling differences in environment dynamics and to further parametrize a policy. Similarly, \citet{galashov2019meta} propose a probabilistic framework for sequential decision-making that they instantiate with NPs for meta-learning. In contrast to this line of works, the novelty of our work is not about a generalizable solution to distributions of RL tasks, but rather about a generalizable method for making modeling with cognitive models practical. This is an important distinction because cognitive models are not necessarily compatible with the RL formalism --- even when they are, they are based on computational rationality, and specifically tailored to account for cognitive limitations. Adapting these limitations to existing frameworks, such as HiP-MDPs, is not trivial and necessarily requires manual effort.

Our work also connects to a line of research studying inference for decision making agents in the context of probabilistic programming. However, most of the approaches make restricting assumptions either regarding the behavior generative processes of the users or the inference objectives and could be applied only for very limited types of problems. For instance, \cite{zhi2020online} consider online inference of boundedly-rational agents but their approach can be applied only in discrete and deterministic environments to capture only agent goals. Furthermore, their solution assumes that the agents start planning their policy from scratch during interaction --- in practical interactive settings, humans might already have a partial or complete plans at the beginning of interaction. On the other hand, many other works, such as by \cite{seaman2018nested}, assume that the likelihood for the generative process $p(\pi \mid \theta)$ can be evaluated for MCMC, which is often an unrealistic assumption with advanced cognitive models.




Many computational approaches motivated by cognitive science share parallels with our objectives. For instance, computational rationality \citep{lewis2014computational, gershman2015computational} and Theory of Mind (ToM) \citep{premack1978does} have motivated numerous computational approaches such as Bayesian ToM \citep{baker2011bayesian}, Machine Theory of Mind \citep{rabinowitz2018machine} and the Menu Search model \citep{chen2015emergence} for modeling human behaviors. Furthermore, \citet{peltola2019interactive} utilize ToM for modeling users with their own models of the interactive system in bandit settings. Among others, these models are prime candidates our method can be applied to.

\section{Discussion} \label{disc}

In this work, we have addressed the so-far unaddressed problem of enabling probabilistic user modeling with complex cognitive models in real-time applications. We introduced a meta-learning approach for training widely applicable differentiable surrogates for approximating posterior predictive estimation with cognitive models. We studied neural process models as example implementations for the surrogates and demonstrated comparable modeling performance to likelihood-free inference with computational cost suitable for online applications. We also showed that the proposed solution allows AI-assistants to utilize cognitive user models computationally feasibly, for instance in a previously studied menu-search task. In a larger scale, the solution not only removes a computational bottleneck currently hindering incorporation of users into probabilistic programming models, but also enables real-time user modeling in various applications where they currently are not possible within usually available computational budgets.

We also demonstrated how the effects of model misspecification in cognitive models can be mitigated in the surrogates by incorporating observed user data in the training. Importantly, we observed that our approach provided robustness against action noise while adapting to the true population as more behavior data became available. Based on these observations, we conclude that the proposed solution can be particularly useful in application domains where user data are limited or behavioral user models can be slightly misspecified, although future studies are still required in settings where the misspecification is caused by more systematic biases.

It is crucial to note that amortization for probabilistic user modeling with cognitive models, as detailed in Section~\ref{dum}, has not been previously widely studied. Apart from LFI, which is computationally intractable for our problems, we are not aware of any solutions which could act as either relevant baselines or alternatives to the proposed approach. Specifically, all the experimented alternative surrogate implementations, such as MAML and transformers, are not fully consistent with the probabilistic nature of the problem, limiting their applicability in practice. We further note that also neural processes feature several compromises in comparison to a fully Bayesian setting with cognitive models: although supporting posterior predictive estimation, they cannot be directly adapted for Bayesian inference in an explicit, predefined parameter space and they do not necessarily follow all constraints coming from the known structure of the behavioral model due to amortization. Future research should adapt and develop alternative surrogate solutions to address these drawbacks.

Interesting avenues for future research include utilizing the surrogates in full probabilistic programming pipelines, although we hypothesize that this should already be possible within certain limits with our approach. Other attractive extensions could consider alternative surrogate architectures to handle, for instance, non-stationarity in cognitive models and settings with multiple data modalities. Regarding ethical considerations, user modeling has always been a double-edged tool and can potentially be abused to serve other interests than those of users --- this should be taken carefully into account in all of its applications. As a generic tool to mitigate some of these issues, we recommend combining user systems with privacy preservation with differential privacy.

\begin{acknowledgements} 
    We would like to thank Sebastiaan De Peuter, Pierre-Alexandre Murena and Sammie Katt for their valuable advice and feedback. This work was supported by the Academy of Finland (Flagship programme: Finnish Center for Artificial Intelligence FCAI and decision 345604) Humane-AI-NET and ELISE Networks of Excellence Centres (EU Horizon: 2020 grant agreements 952026 and 951847), and UKRI Turing AI World-Leading Researcher Fellowship (EP/W002973/1). We also acknowledge the computational resources provided by the Aalto Science-IT Project from Computer Science IT.
    
\end{acknowledgements}

\bibliography{hamalainen_309}

\onecolumn 

\appendix
\section{Experiment 1 details}

\paragraph{Setting.} The first experiment scenario considers modeling Monte Carlo Tree Search (MCTS) \citep{browne2012survey} agents in a simple $10 \times 10$ gridworld environment.  The gridworld environment is defined as a partially observable Markov decision process (POMDP) with deterministic transition dynamics. The state and action spaces are shared across the all agents, including the transition function; the reward and observation functions are individual for each agent. 

The state space $\mathcal{S} = \{1,\dots,10\}^2$ captures all possible agent locations, i.e., grid states. The action space is defined as $\mathcal{A} = \{ \text{up}, \text{down}, \text{left}, \text{right} \}$, the transition function corresponds to transitioning to adjacent grid states in relation to the agent's current location according to the selected action. Actions attempting to relocate agents out of the grid do not cause state transitions.

The agents are fully described by their generative process $\pi \sim p(\pi \mid \theta)$ (MCTS) and parameters $\theta \sim p(\theta)$. Their respective distributions are described in Table~\ref{tab:ex11}. In addition to the reward function, determining one positive and one negative reward state, the prior determines the agent memory, observation function, and MCTS planning-tree depth. The agent memory is defined as a binary parameter determining if the agents utilize the planning tree from previous time steps for subsequent planning instead of starting from scratch. The agents are divided into two subspecies depending on their observation function. Specifically, the complete gridworld is fully observable for half of the agent population, while the other half cannot observe reward states that are located beyond planning-tree depth. In practice, such agents avoid exploring the environment if the positive reward is not directly observed.

\begin{table}[h]
    \centering
    \caption{Uniform prior on user model parameters for the user population in gridworld experiment.}
    \begin{tabular}{l|l}
        \textbf{User Parameter} & \textbf{Distribution} \\ \hline 
         Reward States ($x, y$) &  $\mathcal{U}\{1,\dots,10\}$ \\
         Memory                 &  $\mathcal{U}\{0,1\}$ \\
         Observation function   &  $\mathcal{U}\{0,1\}$ \\
         Tree Depth             &  $\mathcal{U}\{5,\dots,10\}$ \\
    \end{tabular}
    \label{tab:ex11}
\end{table}

Here the task is to model individual users sampled from the population $\theta \sim p(\theta)$. The users generate $n \sim \mathcal{U}\{1,\dots,8\}$ trajectories of length $10$, each collected from one episode in the environment. A new initial user state, $x,y \sim \mathcal{U}\{1,\dots,10\}$, is sampled at each episode. The trajectories are then partitioned into context and target data for NP training by randomly selecting and then truncating one target trajectory  at length $l \sim \mathcal{U}\{1, \dots 9\}$. The beginning of the truncated trajectory, together with the other trajectories, forms the context dataset while the remaining half is held-out as modeling target. Based on the given context data, we evaluate prediction accuracy on the held-out target datasests over tasks. Unlike in the rest of our experiments, each method is here trained with data from $10000$ users.

\paragraph{Implementation.}

All NP models are implemented on the basis of the \verb|NeuralProcesses.jl| library, a Julia variant of the neural process library of Dubois et al., (2020). The MCTS agents are implemented by utilizing the \verb|POMDPs.jl| (Egorov et al., 2017) library. All NP models are meta-trained on a single GPU (NVIDIA Quadro P2200). The code used to produce all the results in this paper can be found at \url{https://github.com/hamalajaa/DifferentiableUserModels}.

The base-architecture is shared among all NP models (summarized in Table~\ref{tab:ex12}). We implement the MAML as first-order MAML \citep{finn2017model} to reduce test-time computation. The Reptile, MAML and transformer details are included in Tables~\ref{tab:ex13}~and~\ref{tab:ex14} (Remark: here, the 'fifth' action corresponds to 'staying still' when the positive reward is found.)

\begin{table}[h]
    \centering
    \caption{Base-architecture shared by all NP models in experiment 1.}
    \begin{tabular}{l r|l r}
        \textbf{Encoder} &  & \textbf{Decoder} & \\ \hline 
         Number of layers    & $6$      & Number of layers    & $6$         \\
         Activations       & Leaky ReLU & Activations         & Leaky ReLU  \\
         Hidden dimensions   & $128$    & Hidden dimensions   & $128$       \\ 
         Latent distribution & Gaussian & Output distribution & Categorical \\
         Input dimensions    & $2$      & Output dimensions   & $5$  
    \end{tabular}
    \label{tab:ex12}
\end{table}

\begin{table}[]
    \centering
    \caption{The architecture shared by Reptile, MAML, and transformer in experiment 1.}
    \begin{tabular}{l|l}
         \textbf{Architecture}         & \\ \hline
         Layers & Transformer + $6$ MLP \\
         Activations & ReLU (+ softmax) \\
         Hidden dimensions   & $128$    \\ 
         Input dimensions    & $2$ \\
         Output dimensions   & $5$ 
    \end{tabular}
    \label{tab:ex13}
\end{table}

\begin{table}[]
    \centering
    \caption{Reptile, MAML, and transformer training and evaluation details in experiment 1.}
    \begin{tabular}{l||l|l|l}
         \textbf{Training} & \textbf{Reptile} & \textbf{MAML} & \textbf{Transformer} \\ \hline
         Optimizer and learning rate & - & - & Adam, $5 \cdot 10^{-4}$  \\
         Meta optimizer & Adam, $5 \cdot 10^{-3}$ & Gradient descent, $5 \cdot 10^{-3}$ & - \\
         Batch optimizer & Gradient descent, $5 \cdot 10^{-3}$ & Gradient descent, $5 \cdot 10^{-3}$ & - \\
         Loss & Cross entropy & Cross entropy & Cross entropy   \\ \hline
         \textbf{Evaluation} & & & \\ \hline
         Optimizer and learning rate & Gradient descent, $0.01$ & Gradient descent, $0.01$ & - \\
         N of gradient steps & $32$ & $32$ & -
    \end{tabular}
    \label{tab:ex14}
\end{table}

\section{Experiment 2 details}

\paragraph{Setting.} The second experiment is based on the Menu Search model of \citet{kangasraasio2019parameter}. The Menu Search model is a cognitive model describing human search behavior in terms of eye movements (saccades) when searching for a target item in a computer dropdown menu. The model simulates user behaviors as a result of optimization with RL given their cognitive constraints. In this experiment, we implement the users as deep Q-learning agents to reduce data generation costs.

Formally, the environment is specified as a POMDP where states contain information about (1) the user's current knowledge about the menu elements, (2) the current gaze focus, and (3) whether or not the user has closed the menu (i.e., quit). We consider a menu of eight elements, where each element is described with its semantic relevance and length given the user's target element. At each step, the user can either fixate on a menu element or quit the scenario. Fixating on a menu element has a chance to reveal the given element while also having a chance to reveal the adjacent elements via peripheral vision. If user action results in revealing the target element, the element is automatically selected, and a significant positive reward is given.

The target word is not present in $10\%$ of the generated menus. If the user recognizes that the target element is not present and quits the menu, a large reward is emitted. For each action, the user is otherwise given a small negative reward based on the duration of the action specified by cognitive parameters presented in Table~\ref{tab:ex2} as priors. When entering a menu for the first time, there is a chance, $p_{rec}$, that the user recalls the menu, completely revealing the entire menu layout. 

\begin{table}[h]
    \centering
    \caption{Distributions for user cognitive properties used in the second experiment.}
    \begin{tabular}{r c|l}
         \textbf{User Parameter} & & \textbf{Distribution}  \\ \hline
         Menu recall probability & $p_{rec}$ & $\text{\textit{Beta}}(3.0, 1.35)$ \\
         Eye fixation duration   & $f_{dur}$ & $\mathcal{N}(3.0, 1.0)$\\
         Target item selection delay & $d_{sel}$ & $\mathcal{N}(0.3, 0.3)$
    \end{tabular}
    \label{tab:ex2}
\end{table}

Similarly as in the first experiment, each user completes $n \sim \mathcal{U}\{1,\dots,8\}$ search tasks with independently sampled menus and target elements constructing $n$ trajectories. Similarly, as in the first experiment, we truncate one of the trajectories to form global and local contexts and the prediction target for ANP training. The ANP, Reptile, MAML, and transformer architectures follow the design used in the first experiment (except for the input and output dimensions).

\paragraph{Implementation.} The implementation details for ANP, Reptile, MAML, and transformer architectures follow the design used in the first experiment (except for the input and output dimensions).

\section{Experiment 3 details}

The third experiment extends the menu search environment into a relatively realistic AI-assistant scenario. First, the menu environments are scaled to consider two levels of hierarchy: each full menu consists of a main menu whose elements correspond to labels that (1) act as links to sub-menus and (2) summarize the contents of these menus. Secondly, we introduce an AI assistant equipped with the proposed user modeling system. The task of the assistant is to utilize the modeling system to propose sub-menus for the users. Intuitively, a successful assistant should guide the users to menus that are likely to contain the true target to shorten their search time.

\paragraph{Environment.} The hierarchical menu search environment introduces an $8 \times 8$ two-level menu setting. Importantly, the environment behaves otherwise similarly to the original non-hierarchical version, with the exception of introducing a main menu that allows a user to navigate between multiple menus. In addition, we introduce a simple mapping between user observations (semantic relevancies and lengths w.r.t. the target element) and assistant observations (logical groups). Specifically, each scenario introduces a set of $8$ logical groups $\mathcal{S}_{AI} = \{1,\dots,8\}$ and $4$ semantic relevance groups $\mathcal{S}_{user} = \{target (1), high (2), medium (3), low (4)\}$ and an independently generated bidirectional mapping between $\mathcal{S}_{AI}$ and $\mathcal{S}_{user}$. The mapping initializes an ordered set of relevancies as $r = \{4,4,4,3,3,2,3,3\}$ and assigns a relevance for each logical group with a randomized circular shift on $r$. The intuition of the mapping is simply to mask the semantic information regarding the target element (via randomization) while allowing a soft prior heuristic for the assistant by conserving semantic similarity between similar logical groups. We similarly mask the item lengths via randomization.

After the mapping between the observation spaces $\mathcal{S}_{AI}$ and $\mathcal{S}_{user}$ is constructed, we sample two logical groups for each sub-menu (such that each group occurs exactly twice in the full menu) and determine a semantic label for the menus summarizing the relevancies of their respective logical groups. The target element is then assigned randomly into one of the sub-menus that includes a logical group with $high$ relevance. The contents for each sub-menu are otherwise determined by mapping the semantic labels of their logical groups into individual items according to the original menu search model specifications. The main menu similarly follows the original specifications --- however, we utilize the semantic labels of the corresponding sub-menus as the relevancies for the main menu elements. At the main menu level, we also replace the item length information with a binary variable denoting if the user has already opened the corresponding sub-menu. Finally, the transition dynamics between the main menu and sub-menus are defined as follows: selecting an element at the main menu -level transitions the user to the corresponding sub-menu, while quitting a sub-menu transitions the environment state back to the main menu. Otherwise, all the transition and reward dynamics follow the original environment specifications.

\paragraph{Assistant.} The hierarchical menu search setting involves a simple search assistant guided by a pre-trained ANP-based user model (user model implementation and training details are described below). In each scenario, the assistant is initially inactive and only activates if the user fails to find the target element from the first sub-menu it explores. When activated, the assistant may suggest and highlight an individual main menu element when the user is at the main menu level. A highlighted main menu element is assumed to attract the attention of the user at its next action and the user's gaze is guided towards the highlighted element. Simultaneously, we assume that the user features some degree of trust towards the assistant's suggestion and the semantic relevance score of the highlighted element is increased by one level. In practice, this allows the user also to reject poor suggestions.

We implement the assistant as a simple rule-based agent that continuously updates the user model as new user actions are observed. We assume that the assistant can track users' gaze locations but that it does not have access to the semantic relevancies of the items. Instead, the assistant updates its estimate on the currently observed (and unobserved) menu elements in terms of the observation space $\mathcal{S}_{AI}$ specified above. When activated, the assistant simulates one user action at fully observed main menu -level conditioned on the observed user search behavior: $a \sim p_\phi(a \mid s, z)$, $z \sim p_\psi(z \mid (\mathbf{s}, \mathbf{a}))$. The main menu element corresponding to the estimated most likely user action is then selected as the assistant's suggestion.

\paragraph{Implementation and training details.} The ANP-based user model is meta-trained on a single GPU. Each user generates $1$ trajectory which is split at length $l \sim \mathcal{U}\{2,\dots,10\}$ into context and target trajectories for ANP training. The base architectures of the ANP, Reptile, MAML and transformer models are identical to the previous experiment. The online prediction times are run on a laptop CPU (Intel Core i7-7700HQ).

\end{document}